\documentclass[11pt,letterpaper]{article}
\usepackage[hyperref]{emnlp2017}
\usepackage{times}
\usepackage{latexsym}
\usepackage{multirow}
\usepackage{graphicx}
\usepackage{enumitem}
\usepackage{booktabs}
\usepackage{amsmath}
\usepackage{subcaption}

\usepackage{balance}
\usepackage{color,soul}
\usepackage{xspace}

\newcommand{\modcce}{\textit{Categorical Cross Ent.}\xspace}
\newcommand{\modmse}{\textit{MSE}\xspace}

\newcommand{\modclass}{\textit{Class Metric}\xspace}

\emnlpfinalcopy

\def\emnlppaperid{***}

\newcommand{\tablefontsize}{\small}%\footnotesize}
\newcommand{\captionfontsize}{}%\footnotesize}

\title{Depression and Self-Harm Risk Assessment in Online Forums}

\author{
{Andrew Yates}$^\dagger$\Thanks{\enspace The first two authors contributed equally to this work.}
\quad {Arman Cohan}$^\ddagger$\footnotemark[1]
\quad {Nazli Goharian}$^\ddagger$\vspace{4pt}\\
$^\dagger$Max Planck Institute for Informatics,\\ Saarland Informatics Campus Saarbruecken, Germany\vspace{4pt}\\
$^\ddagger$Information Retrieval Lab, Department of Computer Science,\\ Georgetown University, Washington  DC, USA\\
{\small \tt ayates@mpi-inf.mpg.de}\\
{\small \tt \{arman,nazli\}@ir.cs.georgetown.edu}
}

\date{}

\begin{document}

\maketitle
\begin{abstract}
Users suffering from mental health conditions often turn to online resources for support, including specialized online support communities or general communities such as Twitter and Reddit.
In this work, we present a neural framework for supporting and studying users in both types of communities. We propose methods for identifying posts in support communities that may indicate a risk of self-harm, and demonstrate that our approach outperforms strong previously proposed methods for identifying such posts. Self-harm is closely related to depression, which makes identifying depressed users on general forums a crucial related task. We introduce a large-scale general forum dataset (``RSDD'') consisting of users with self-reported depression diagnoses matched with control users. We show how our method can be applied to effectively identify depressed users from their use of language alone. We demonstrate that our method outperforms strong baselines on this general forum dataset.
\end{abstract}

\section{Introduction}
Mental health remains a major challenge in public health care. Depression is one of the most common mental disorders and 350 million people are estimated to suffer from depression worldwide \cite{world2010world}. In 2014 an estimated 7\% of all U.S. adults had experienced at least one major depressive disorder \shortcite{depression2015}.
Suicide and self-harm are major related concerns in public mental health. Suicide is one of the leading causes of death \cite{cdc2015}, and each suicide case has major consequences on the physical and emotional well-being of families and on societies in general. Therefore identifying individuals at risk of self-harm and providing support to prevent it remains an important problem \cite{ferrari2014burden}.

Social media is often used by people with mental health problems to express their mental issues and seek support.
This makes social media a
significant resource for studying language related to depression, suicide, and self-harm, as well as understanding the authors' reasons for making such posts, and identifying individuals at risk of harm \cite{coppersmith2014quantifying}.
Depression and suicide are closely related given that depression is the psychiatric diagnosis most commonly associated with suicide.
Research has demonstrated that forums are powerful platforms for self-disclosure and social support seeking around mental health concerns \cite{de2014mental,manikonda2017modeling}. Such support forums are often staffed by moderators who are mental health experts, trained volunteers, or more experienced users whose role is to identify forum posts suggesting that a user is at risk of self-harm and to provide support.

Studies have shown that self expression and social support are beneficial in improving the individual's state of the mind \cite{turner1983social,choudhury2017language} and, thus such communities and interventions are important in suicide prevention. However, there are often thousands of user posts published in such support forums daily, making it difficult to manually identify individuals at risk of self-harm. Additionally, users in acute distress need prompt attention, and any delay in responding to these users could have adverse consequences.
Therefore, identifying individuals at risk of self-harm in such support forums is an important challenge.
Identifying signs of depression in general social media, on the other hand, is also a difficult task that has applications for both better understanding the relationship between mental health and language use and for monitoring a specific user's state (e.g., in the context of monitoring a user's response to clinical care).
In this work we propose and evaluate a framework for performing self-harm risk assessment and for identifying depression in online forums.

We present a general neural network architecture for combining posts into a representation of a user's activity that is used to classify the user.
To address the challenge of depression risk assessment over the general forums, we introduce a large-scale novel Reddit dataset that is substantially larger than the existing data and has a much more realistic number of control users. The dataset contains over 9,000 users with self-reported depression diagnoses matched with over 107,000 control users. We apply our approach to \textit{(1)} identify the users with depression on a general forum like Reddit, and to
\textit{(2)} estimate the risk of self-harm indicated by posts in a more specific mental-health support forum.
 Our methods perform significantly better on both datasets than strong existing methods, demonstrating that our approach can be used both to identify depressed users and to estimate the risk of self-harm posed by individual posts.

\section{Related Work}
\label{sec:related}
There is a growing body of related work analyzing mental health-related discourse and language usage in social media to better discover and understand mental health related concerns \cite{resnik2013using,dechoudhury2013predicting,coppersmith2014measuring,coppersmith2014quantifying,mitchell2015quantifying,tsugawa2015recognizing,coppersmith2015adhd,althoff2016large,mowery2016towards,benton-mitchell-hovy:2017:EACLlong}.
To investigate NLP methods for identifying depression and PTSD users on Twitter, a shared task \cite{coppersmith2015clpsych} at the 2nd Computational Linguistics and Clinical Psychology Workshop (CLPsych 2015) was introduced where the participants evaluated their methods on a dataset of about 1800 Twitter users. Other work has used data from approximately 900 Reddit.com users to support self-reported diagnosis detection \cite{losada2016test}.
Previous work identifying depression and other mental health problems, including the methods participating in CLPsych 2015 (e.g. works by \citet{ resnik2015university} and \citet{preoctiucpietro-EtAl:2015:CLPsych1}) heavily rely on utilizing features such as LIWC \cite{pennebaker2015development}, topic modeling, manual lexicons, or other domain-dependent application-specific features. Aside from the effort required to design effective features, these approaches usually model the problem with respect to the selected features and ignore other indicators and signals that can improve prediction. In contrast, our model only relies on text and is not dependent on any external or domain-specific features. Existing self-reported diagnosis detection datasets
contain a limited number of both control users and diagnosed users. In contrast to this, we construct a new dataset with over 9,000 depressed users matched with a realistic number of control users.

In addition to general studies addressing mental health, related work has also specifically studied suicide and self-harm through social media \cite{jashinsky2014tracking,thompson2014,gunn2015twitter,de2016discovering,coppersmith2016exploratory}. Recently, CLPsych 2016 \cite{CLPsych:2016} investigated approaches for detecting the self-harm risk of mental health forum posts \cite{milne2016clpsych}. Most related work in this area uses variations of linear classifiers with some sort of feature engineering; successful methods have employed: a combination of sparse (bag-of-words) and dense (doc2vec)
representation of the target forum posts \cite{kim-EtAl:2016:CLPsych}, a stack of feature-rich Random Forest and linear Support Vector Machine (SVM) \cite{malmasi-zampieri-dras:2016:CLPsych}, an RBF SVM classifier utilizing similar sets of features \cite{brew:2016:CLPsych}, and various contextual and psycholinguistic features \cite{cohan2015triaging,cohan2017triaging}.
In contrast to the above works, our model does not use any general or domain specific feature engineering; it learns appropriate representations of documents by considering only their textual content.

Our proposed models consist of a shared architecture based on a CNN, a merge layer, model-specific loss functions, and an output layer (as we will describe in \S \ref{sec:method}).
While our model shares similarities with CNN-based models in prior work \cite{Kal:2014,Kim:2014,xiao2016efficient}, it focuses on learning representations of user's posts and combining the post representations into an overall representation of the user's activity.
In the case of self-harm risk assessment, we experiment with several loss functions to determine
whether considering the ordinal nature of self-harm risk labels (i.e., green, amber, red, and crisis)
can improve performance. Evaluation results suggest that the model variant using this loss function is more robust than our other variants.

\section{Data}
\label{sec:data}
\subsection{Depression dataset construction.} We created a new dataset to support the task of identifying forum users with self-reported depression diagnoses.
The Reddit Self-reported Depression Diagnosis (RSDD) dataset
 was created by annotating users from a publicly-available Reddit dataset\footnote{\url{https://files.pushshift.io/reddit/}}. Users to annotate were selected by identifying all users who made a post between January 2006 and October 2016 matching a high-precision diagnosis pattern.\footnote{e.g., ``I was just diagnosed with depression.''} Users with fewer than 100 posts made before their diagnosis post were discarded. Each of the remaining diagnosis posts was then viewed by three layperson annotators to decide whether the user was claiming to have been diagnosed with depression; the most common false positives included hypotheticals (e.g., ``if I was diagnosed with depression''), negations (e.g., ``it's not like I've been diagnosed with depression''), and quotes (e.g., ``my brother announced `I was just diagnosed with depression'\,''). Only users with at least two positive annotations were included in the final group of diagnosed users.

A pool of potential control users was identified by selecting only those users who had \textit{(1)} never posted in a subreddit related to mental health, and \textit{(2)} never used a term related to depression or mental health. These restrictions minimize the likelihood that users with depression are included in the control group.
In order to prevent the diagnosed users from being easily identified by the usage of specific keywords that are
never used by the control users, we removed all posts by diagnosed users that met either one of the aforementioned
conditions (i.e., that was posted in a mental health subreddit or included a depression term).

For each diagnosed user and potential control user, we calculated the probability that the user would post in each subreddit (while ignoring diagnosed users' posts made to mental health subreddits). Each diagnosed user was then greedily matched with the 12 control users who had the smallest Hellinger distance between the diagnosed user's and the control user's subreddit post probability distributions, excluding control users with 10\% more or fewer posts than the diagnosed user. This matching approach ensures that diagnosed users are matched with control users who are interested in similar subreddits and have similar activity levels, preventing biases based on the subreddits users are involved in or based on how active the users are on Reddit.
This yielded a dataset containing 9,210 diagnosed users and 107,274 control users. On average each user in the dataset
has 969 posts (median 646). The mean post length is 148 tokens (median 74).

The Reddit Self-reported Depression Diagnosis (RSDD) dataset differs from prior work creating self-reported diagnoses datasets in several ways: it is an order of magnitude larger, posts were annotated to confirm that they contained claims of a diagnosis, and a realistic number of control users were matched with each diagnosed user.
The lists of terms related to mental health, subreddits related to mental health, high-precision depression diagnosis patterns, and further information are available\footnote{\url{http://ir.cs.georgetown.edu/data/reddit\_depression/}}.
We note that this dataset has some (inevitable) caveats: \textit{(i)} the method only captures a subpopulation of depressed people (i.e. those with self-reported diagnosis), \textit{(ii)} Reddit users may not be a representative sample of the population as a whole, and \textit{(iii)} there is no way to verify whether the users with self-reported diagnoses are truthful.

\subsection{Self-harm assessment.} For self-harm risk assessment we use data from mental health forum posts from ReachOut.com, which is a successful Australian support forum for young people. In addition to providing peer-support, ReachOut moderators and trained volunteers monitor and participate in the forum discussions. The NAACL 2016 Computational Linguistics and Clinical Psychology Workshop \cite{CLPsych:2016} released a Triage dataset containing 65,024 forum posts from ReachOut, with annotations for 1,227 posts indicating the author's risk of self-harm \cite{milne2016clpsych}. The annotations consist of one of four labels: green (indicating no action is required from ReachOut's moderators), amber (non-urgent attention is required), red (urgent attention is required), and crisis (a risk that requires immediate attention).

\subsection{Ethical concerns.}
Social media data are often sensitive, and even more so when the data are related to mental health.
Privacy concerns and the risk to the individuals in the data should always be considered \cite{hovy2016social,vsuster2017short,benton2017ethical}. We note that the risks associated with the data used in this work are minimal. This assessment is supported by previous work on the ReachOut dataset \cite{milne2016clpsych}, on Twitter data \cite{coppersmith2015clpsych}, and on other Reddit data \cite{losada2016test}. The RSDD dataset contains only publicly available Reddit posts. Annotators were shown only anonymized posts and agreed to make no attempts to deanonymize or contact them.
The RSDD dataset will only be made available to researchers who agree to follow ethical guidelines,
which include requirements not to contact or attempt to deanonymize any of the users.
Additionally, for the ReachOut forum data that was explicitly related to mental health, the forum's rules require the users to stay anonymous; moderators actively redact any user identifying information.

\begin{figure}[tb]
\begin{center}
\includegraphics[scale=0.34]{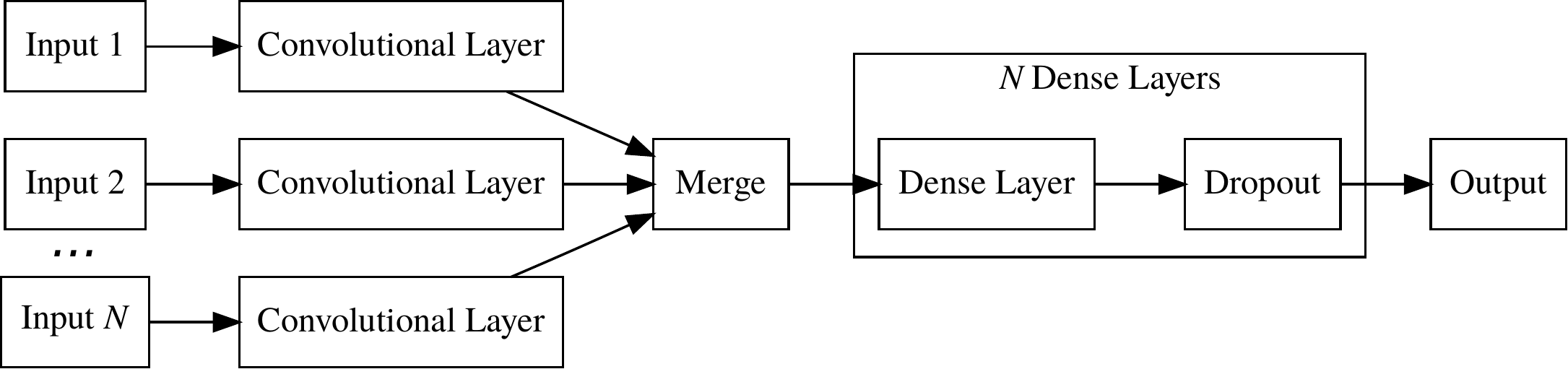}
\caption{\captionfontsize
The general neural network architecture shared among our user and post classification models. Each input (e.g., each of a user's posts) is processed by a convolutional network and merged to create a vector representation of the user's activity. This vector representation is passed through one or more dense layers followed by an output layer that performs classification. The type of input received, merge operation, and output layer vary with the specific model.}
\label{fig:arch}
\end{center}
\end{figure}

\section{Methodology}
\label{sec:method}
We describe a general neural network architecture for performing text classification over multiple input texts.
We propose models based on this architecture for performing two tasks in the social media and mental health domains that we call \textit{self-harm risk classification} and \textit{detecting depression}.
The task of self-harm risk classification is estimating a user's current self-harm risk given the user's post on a mental health support forum and the previous posts in the thread. The task of detecting depressions in users is identifying Reddit users with self-reported depression diagnoses given the users' post histories (excluding posts containing mental health keywords or posted in subreddits related to mental health).

While both tasks are focused on predicting a user's mental health status, they differ in both the type of classification performed (i.e., estimating severity on a four point scale vs. boolean classification) and in the amount of data available.
Our general architecture is based on a two step process: \textit{(1)} identifying relevant features in each input text, and \textit{(2)} combining the features observed in the model's inputs to classify the user.

\begin{figure}[tb]
\begin{center}
\includegraphics[scale=0.45]{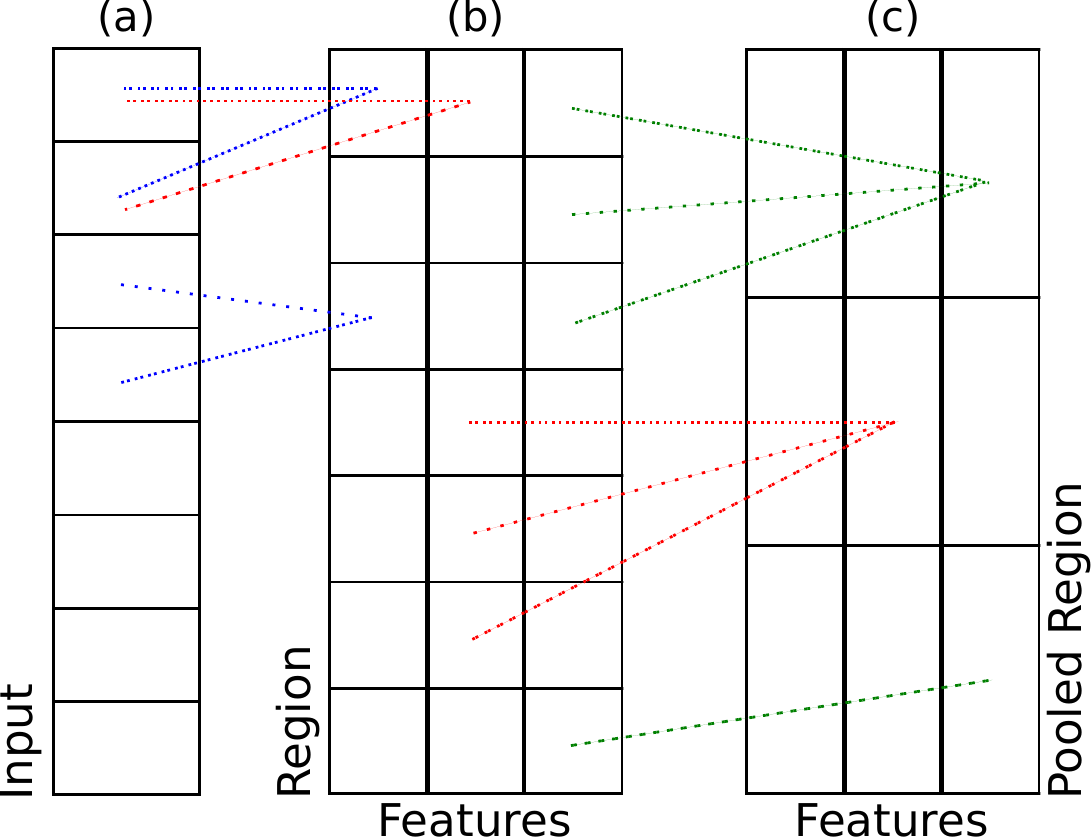}
\caption{\captionfontsize
The convolutional network component of our architecture. A convolutional layer takes a series of terms as input \textit{(a)} and applies $l$ filters to a $k$-term sliding window to derive feature values for each window or region \textit{(b)}; $k=2$ and $l=3$ shown here. A max pooling layer considers non-overlapping region sequences of length $n$ \textit{(b)} and keeps the highest feature value for the sequence \textit{(c)}; $n=3$ shown here.}
\label{fig:conv}
\end{center}
\end{figure}

\subsection{Shared Architecture}
Our proposed models share a common architecture that takes one or more posts as input,
processes the posts using a convolutional layer to identify features present in sliding windows of text,
merges the features identified into a vector representation of the user's activity,
and uses a series of dense layers to perform classification on the merged vector representation.
The type of merging performed and the output layers are properties of the model variant,
which we describe in detail in the following section.
Convolutional networks have commonly been applied to the task of text classification, such as by \newcite{Kim:2014}.
We use categorical cross-entropy as a loss function with both methods, but also experiment with
other loss functions when performing severity classification.

First, the model takes one or more posts as input and processes each post with a convolutional network
containing a convolutional layer and a pooling layer. This process is illustrated with a max pooling layer in Figure \ref{fig:conv}.
The convolutional layer applies filters to a sliding window of $k$ terms \textit{(a)} and outputs a feature value
for each sliding window region and each filter \textit{(b)}. The same filters are applied to each window;
each filter can be viewed as a feature detector and the overall process can be conceptualized
as looking for windows of terms that contain specific features. The features are not specified a priori through
feature engineering, but instead are learned automatically when the model is trained. After identifying the features
present in each region (i.e., sliding window), a max pooling layer considers non-overlapping regions of length $n$ and keeps the highest feature value for each region \textit{(c)}.
This step eliminates the regions (i.e., sliding windows) that do not contain useful features, which reduces the size of the convolutional network's output.
The same convolutional network is applied to each
input post, meaning that the model learns to look for the same set of features in each.

After each input post has been processed by a convolutional network, the output of each convolutional
network is merged to create a representation of the user's
activity across all input posts. This representation is processed by one or more dense layers (i.e., fully
connected layers) with dropout \cite{srivastava2014dropout}
before being processed by a final output layer to perform classification.
The type of output layer is dependent on the model variant.
Our shared model architecture is illustrated in Figure \ref{fig:arch}.
The architecture's hyperparameters (e.g.,
the sliding window size $k$, the number of convolutional filters used, and type of pooling)
also vary among models and are described in \S\ref{sec:exp}.
Both the convolutional and dense layers use ReLU activations \cite{icml2010_NairH10} in all model variants.

\begin{table*}[tb]
\begin{center}
\small
\begin{tabular}{@{}llcccccr@{}}
\toprule
\multicolumn{2}{c}{\multirow{2}{*}{Method}} & \multicolumn{3}{c}{Convolution} &
\multirow{2}{*}{Dense Layers} & \multirow{2}{*}{Dropout} & \multirow{2}{*}{Class Balance}  \\
\cmidrule(lr){3-5}
 & & Size  & Filters & Pool Len. &  &  &   \\
\midrule
Reddit & Cat. Cross Ent. & 3 & 25 & all (avg) & 1 w/ 50 dims & 0.0 & Sampled \\
\midrule
ReachOut & Cat. Cross Ent. & 3 & 150 & 3 (max) & 2 w/ 250 dims & 0.3 & Weighted \\
& MSE & 3 & 100 & 3 (max) & 2 w/ 250 dims & 0.5 & Sampled \\
& Class Metric & 3  & 100 & 3 (max) & 2 w/ 150 dims & 0.3 & Sampled \\
\bottomrule
\end{tabular}
\caption{The hyperparameters used by each model.}
\label{table:params}
\end{center}
\end{table*}

\subsection{Models}

\subsubsection{Depression detection}
Our model for depression detection takes a user's posts as input and processes each post
with a convolutional network. Each convolutional network performs average pooling to produce its output.
These post representations are then merged with a second convolutional layer to create a user representation;
we found this approach led to more stable performance than using a second average pooling or max pooling layer.
The user representation created by the merge step is then passed to one or more dense layers before being passed
to a dense output layer with a softmax activation function to perform classification.
The number of dense layers used is a hyperparameter described in \S \ref{sec:exp}.
Categorical cross-entropy is used as the model's loss function.

\subsubsection{Self-harm risk assessment}
Our model for self-harm risk classification takes two inputs: the target post being classified
and the prior posts (if any) in the target post's thread. The prior posts provide context and are thus useful
for estimating the risk of self-harm present in the target post.
The two inputs are both processed by a convolutional network as in user-level classification, but in this case
the convolutional network's outputs correspond to a representation of the target post and to a representation
of the target post's context (i.e., the prior posts in the thread). Given that these two outputs represent
different aspects, they are merged by concatenating them together.
This merged representation is then passed to one or more dense layers and to an
output layer; the type of output layer depends on the loss function used.
There are four self-harm risk assessment model variants in total:

\modcce uses an output layer with a softmax activation function,
and categorical cross-entropy as its loss function. This mirrors the output layer and loss function
used in the user level classification model.

\modmse uses an output layer with a linear activation function, and mean squared error as its loss function.
The model's output is thus a single value;
to perform classification, this output value is rounded to the nearest integer in the interval $[0, t - 1]$,
where $t$ is the number of target classes.

The final two loss functions perform metric learning rather than performing classification directly. They learn representations of a user's activity and of the four self-harm risk severity labels; classification is performed by comparing the euclidean distance between a representation of a user's activity (produced by the final layer) and each of the four severity label representations.

\modclass: Let $d$ be the size of the output layer and $X$ be the layer's $d$-dimensional output. \modclass learns a $d$-dimensional representation of each class $C_i$ such that $||X - C_i||_2$ is minimized for the correct class $i$; this is accomplished with the loss function:
$$ L_{i,p,n} = \max \big(0, ||X_i - C_p||_2 - ||X_i - C_n||_2 + \alpha \big) $$

\noindent where $C_p$ is the correct (i.e., positive) class for $X_i$, $C_n$ is a randomly chosen incorrect (i.e., negative) class, and
$\alpha$ is a constant to enforce a minimum margin between classes. Classification is performed by
computing the similarity between $X_i$ and each class $C_j$.

\modclass \textit{(Ordinal)} extends \modclass to enforce a margin between ordinal classes as a function of the distance between classes. Given a ranked list of classes such that more similar classes have closer rankings, that is $\forall i$ $sim(C_i, C_{i \pm 1}) > sim(C_i, C_{i \pm 2})$, we incorporate the class distance into the margin such that more distant incorrect class labels must be further away from the correct class label in the metric space. The loss function becomes
\begin{multline*} L_{i,p,n} =  \max \big(0, \;\; ||X_i - C_p||_2 \;\; - \\
||X_i - C_n||_2 + \alpha|p-n| \big)
\end{multline*}
\noindent where $|p-n|$ causes the margin to scale with the distance between classes $p$ and $n$.

\section{Experiments}
\label{sec:exp}
In this section, we describe the model hyperparameters used and present our results on the depression detection
and self-harm risk assessment tasks.
To facilitate reproducibility we provide our code and will provide the Reddit depression dataset
to researchers who sign a data usage agreement\footnote{\url{http://ir.cs.georgetown.edu/data/reddit\_depression/}}.

\subsection{Experimental setup.}
\label{sec:setup}
The hyperparameters used with our models are shown in Table~\ref{table:params}.
The severity risk assessment models' hyperparameters were chosen using 10-fold cross validation on the 947 ReachOut training posts, with 15\% of each fold used as validation data.
The depression identification model's hyperparameters were chosen using the Reddit validation set.
The depression identification model's second convolutional layer (i.e., the layer used to merge post representations)
used filters of length 15, a stride of length 15, and the same number of filters as the first convolutional layer.
All models were trained using stochastic gradient descent with the Adam optimizer \cite{adam}.
The hyperparameters that varied across models are shown in Table \ref{table:params}. The convolution size, number of convolutional filters, pooling type, pooling length, and number of dense layers was similar across all post models. Class balancing was performed with \modcce by weighting classes inversely proportional to their frequencies, whereas sampling an equal number of instances for each class worked best with the other methods.

\paragraph{Addressing limited data.}
The post classification models' input consists of skip-thought vectors \cite{Kiros15};
each vector used is a 7200-dimensional
representation of a sentence. Thus, the convolutional windows used for post classification are over sentences
rather than over terms. This input representation was chosen to mitigate the effects of the ReachOut dataset's
relatively small size. The skip-thought vectors
were generated from the the ReachOut forum dataset by sequentially splitting the posts in the training set into sentences, tokenizing them, and training skip-thoughts using Kiros et al.'s implementation with the default parameters. Sentence boundary detection was performed using the Punkt sentence tokenizer \cite{punkt} available in NLTK \cite{nltk}. These 2400-dimensional forum post skip-thought vectors were concatenated with the 4800-dimensional book corpus skip-thought vectors available from \citeauthor{Kiros15}. Experiments on the training set indicated that using only the ReachOut skip-thought vectors slightly decreased performance, while using only the book corpus skip-thought vectors substantially decreased performance.
As input the post models received the last 20 sentences in each target post and the last 20 sentences in the thread prior to the target post; any prior sentences are ignored.

\subsection{Depression detection.}
The data used for depression detection was described in \S \ref{sec:data}.
As baselines we compare our model against the FastText classifier \cite{joulin2016bag}
and MNB and SVM classifiers \cite{wang2012baselines} using features from prior work.
We tune FastText's hyperparameters on the validation set. Specifically, we consider a maximum
n-gram size $\in [1,2,3,4,5]$, an embedding size $\in [50,100, 150]$, and a learning rate
$\in [0.05, 0.1, 0.25, 0.5]$ as suggested in the documentation.
We consider two sets of features for the MNB and SVM classifiers. The first set of features is the post content itself represented as sparse bag of words features (\textit{BoW baselines}). The second set of features (\textit{feature-rich baselines}) comprises a large set of features including bag of words features encoded as sparse weighted vectors, external psycholinguistic features captured by LIWC\footnote{\url{http://liwc.wpengine.com/}} \shortcite{pennebaker2015development}, and emotion lexicon features \cite{staiano2014depechemood}. Since our problem is identifying depression among users, psycholinguistic signals and emotional attributes in the text are potentially important features for the task. These features (as described in \S \ref{sec:related}) have been also previously used by successful methods in the Twitter self-reported diagnosis detection task \cite{coppersmith2015clpsych}. Thus, we argue that these are strong baselines for our self-reported diagnosis detection task. We apply count based and TF-IDF based feature weighting for bag of words features. We perform standard preprocessing by removing stopwords and lowercasing the input text.\footnote{During experimentation, we found TF-IDF sparse feature weighting to be superior than other weighting schemes. Additional features such as LDA topics and $\chi^2$ feature selection did not result in any further improvements.}

The data is split into training, validation, and testing datasets each containing approximately 3,000 diagnosed users and their matched control users. The validation set is used for tuning development and hyperparameter tuning of our models and the baselines. The reported results are on the test set.
The depression detection models' input consisted of raw terms encoded as one-hot vectors.
We used an input layer to learn 50-dimensional representation of the terms.
For each target user, the CNN received up to $n_{post}$ posts containing up to $n_{term}$ terms.
In this section we present results for two values of $n_{post}$.
The earliest post approach (CNN-E) takes each user's $n_{post}=400$ earliest posts as input.
The random approach (CNN-R) samples $n_{post}=1500$ random posts from each user.
We empirically set $n_{term}=100$ with both approaches.
We later analyze the model's performance as $n_{post}$ and $n_{term}$ vary in \S \ref{sec:maxposts}
and as the post selection strategy varies in \S \ref{sec:selection}.

\begin{table}[tb]
\begin{center}
\tablefontsize
\setlength{\tabcolsep}{6.5pt}
\renewcommand{\arraystretch}{0.8}
\begin{tabular}{@{}lccc@{}}
\toprule
Method & Precision & Recall & F1 \\ \midrule
BoW - MNB           & 0.44 & 0.31 & 0.36 \\
BoW - SVM           & 0.72 & 0.29 & 0.42 \\
Feature-rich - MNB  & 0.69 & 0.32 & 0.44 \\
Feature-rich - SVM  & 0.71 & 0.31 & 0.44 \\
FastText & 0.37 & \textbf{0.70} & 0.49 \\
\midrule
User model - CNN-E          & 0.59 & 0.45 & 0.51 \\
User model - CNN-R          & \textbf{0.75} & 0.57 & \textbf{0.65} \\
\bottomrule
\end{tabular}
\caption{\captionfontsize  Performance identifying depressed users on the Reddit test set.
 The differences between the CNN and baselines are statistically significant (McNemar's test, $p<0.05$).}
\label{table:redditresults}
\end{center}
\end{table}

\begin{table*}[tb]
\begin{center}
\tablefontsize
\renewcommand{\arraystretch}{0.8}
\setlength{\tabcolsep}{5pt}
\begin{tabular}{@{}lccccccc@{}}
\toprule
\multirow{2}{*}{Method} & Non-green & \multicolumn{2}{c}{Flagged} & \multicolumn{2}{c}{Urgent} & \multicolumn{2}{c}{All} \\
\cmidrule(lr){2-2} \cmidrule(lr){3-4} \cmidrule(lr){5-6} \cmidrule(lr){7-8}
 & F1 & F1 & Acc. & F1 & Acc. & F1 & Acc. \\
\midrule
Baseline \cite{milne2016clpsych} & 0.31 & 0.78 & 0.86 & 0.38 & 0.89 & - & - \\ \midrule
\citet{kim-EtAl:2016:CLPsych} & 0.42 & 0.85 & 0.91 & 0.62 & 0.91 & 0.55 & 0.85 \\
\citet{malmasi-zampieri-dras:2016:CLPsych} & 0.42 & 0.87 & 0.91 & 0.64 & 0.93 & 0.55 & 0.83 \\
\citet{brew:2016:CLPsych}         & 0.42 & 0.78 & 0.85 & 0.69 & 0.93 & 0.54 & 0.79 \\
\citet{cohan2015triaging} & 0.41 & 0.81 & 0.87 & 0.67 & 0.92 & 0.53 & 0.80 \\
\midrule
Categorical Cross Ent. & \textbf{0.50} & \textbf{0.89} & \textbf{0.93} & 0.70 & \textbf{0.94} & \textbf{0.61} & \textbf{0.89} \\
MSE & 0.42 & 0.80 & 0.85 & 0.64 & 0.93 & 0.53 & 0.78 \\
Class Metric & 0.46 & 0.79 & 0.84 & 0.70 & \textbf{0.94} & 0.56 & 0.80 \\
Class Metric (Ordinal) & 0.47 & 0.88 & \textbf{0.93} & \textbf{0.72} & 0.93 & 0.59 & 0.87 \\
\bottomrule
\end{tabular}
\caption{\captionfontsize  Self-harm risk assessment performance on the ReachOut CLPsych '16 test set.
Results for the other methods are from \cite{milne2016clpsych}.
Differences in performance between the following pairs are statistically significant (McNemar's test, $p<0.05$):
\modcce and \modclass, \modmse and \modcce, \modmse and \modclass \textit{(Ordinal)}, and \modclass \textit{(Ordinal)} and \modclass.
}
\label{table:results1}
\end{center}
\end{table*}

\begin{table*}[tb]
\begin{center}
\tablefontsize
\renewcommand{\arraystretch}{0.8}
\begin{tabular}{@{}lccccccc@{}}
\toprule
\multirow{2}{*}{Method} & Non-green & \multicolumn{2}{c}{Flagged} & \multicolumn{2}{c}{Urgent} & \multicolumn{2}{c}{All} \\
\cmidrule(lr){2-2} \cmidrule(lr){3-4} \cmidrule(lr){5-6} \cmidrule(lr){7-8}
 & F1 & F1 & Acc. & F1 & Acc. & F1 & Acc. \\
\midrule
Categorical Cross Ent. & 0.54 & 0.87 & 0.89 & 0.69 & 0.91 & 0.63 & 0.80 \\
MSE & \textbf{0.87} & \textbf{0.95} & \textbf{0.96} & \textbf{0.91} & \textbf{0.98} & \textbf{0.89} & \textbf{0.93} \\
Class Metric & 0.73 & 0.90 & 0.91 & 0.81 & 0.94 & 0.78 & 0.86 \\
Class Metric (Ordinal) & 0.85 & \textbf{0.95} & \textbf{0.96} & 0.89 & 0.97 & 0.88 & 0.92 \\
\bottomrule
\end{tabular}
\caption{\captionfontsize Self-harm risk assessment performance on the ReachOut CLPsych '16 training set using 10-fold cross validation. \modcce performs substantially worse than on the test set, while \modmse performs substantially better. \modclass \textit{(Ordinal)} continues to perform well. The difference in performance between the following method pairs are statistically significant (McNemar's test, $p<0.05$): \modcce and \modmse, \modcce and \modclass, \modcce and \modclass \textit{(Ordinal)}, \modmse and \modclass, and \modclass and \modclass \textit{(Ordinal)}.}
\label{table:results1cv}
\end{center}
\end{table*}

\textbf{Results.}
The results of identifying depressed users for our model and baselines are shown in Table \ref{table:redditresults}. Our proposed model outperforms the baselines by a large margin in terms of recall and F1 on the diagnosed users (increases of 41\% and 16\%, respectively), but performs worse in terms of precision.
As described later in the analysis section, the CNN identifies language associated with negative sentiment across a user's posts.

\begin{figure*}[tb]
  \centering
  \begin{subfigure}{0.45\textwidth}
\includegraphics[scale=0.48]{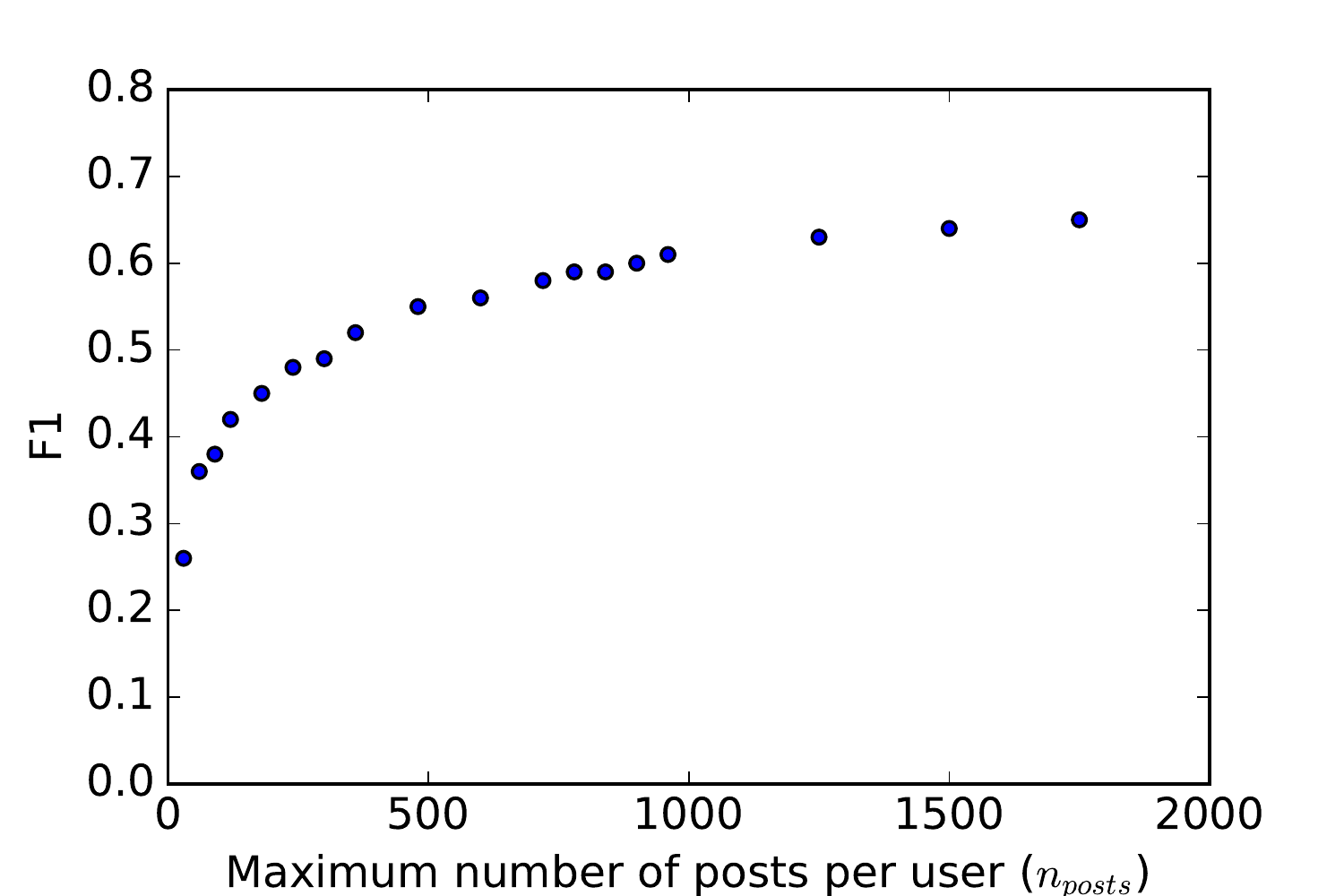}
\caption{}
\label{fig:maxposts}
\end{subfigure}
\begin{subfigure}{0.45\textwidth}
  \includegraphics[scale=0.48]{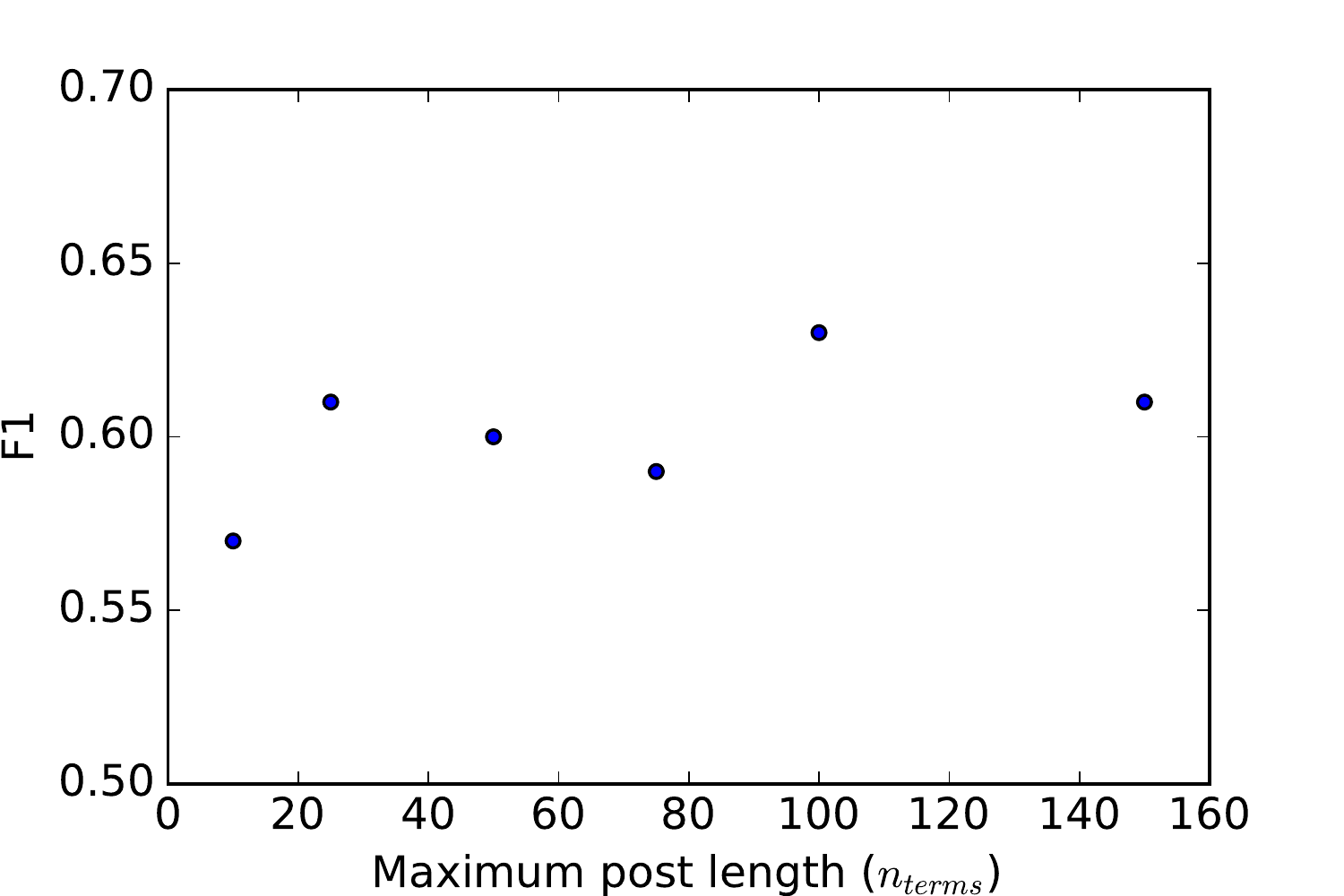}
\caption{}
  \label{fig:maxlen}
\end{subfigure}
\caption{\captionfontsize Sensitivity of the CNN-R model to the parameters $n_{posts}$ (a) and $n_{terms}$ (b) on RSDD's validation set. F1 increases as $n_{posts}$ does (a), but the rate of increase slows as $n_{posts}$ surpasses 1000.
  The trend for $n_{terms}$ is less clear (b), but the highest F1 is achieved at $n_{terms}=100$.
  In Figure (a) the parameter $n_{terms}$ was fixed to 100, and in Figure (b) $n_{posts}$ was fixed to 1500. }\label{fig:model-params}
\end{figure*}

\begin{figure*}
  \centering
  \begin{subfigure}{.45\textwidth}
  \begin{center}
  \includegraphics[scale=0.48]{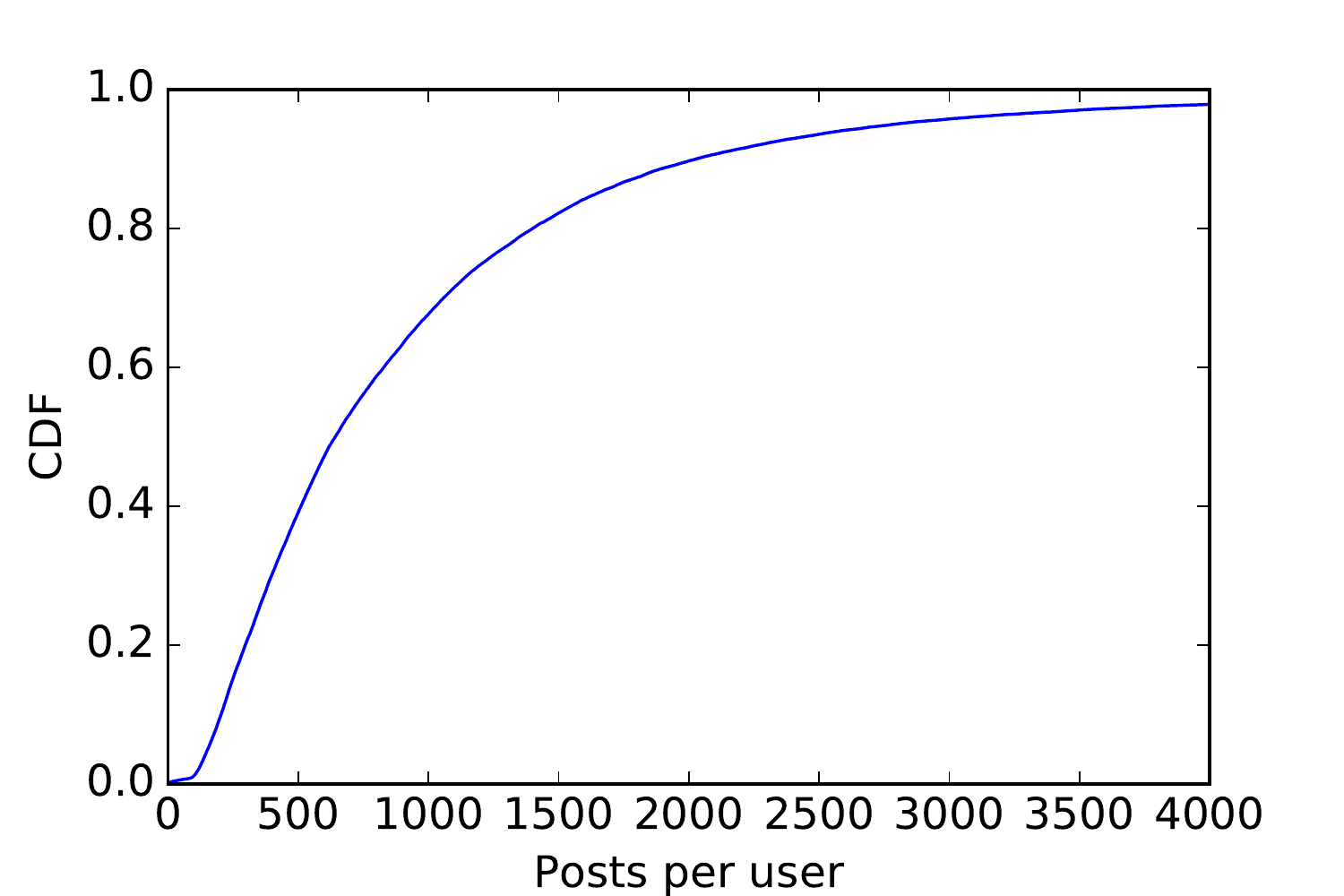}
\caption{}
  \label{fig:postcdf}
  \end{center}
\end{subfigure}
\begin{subfigure}{.45\textwidth}
\begin{center}
\includegraphics[scale=0.48]{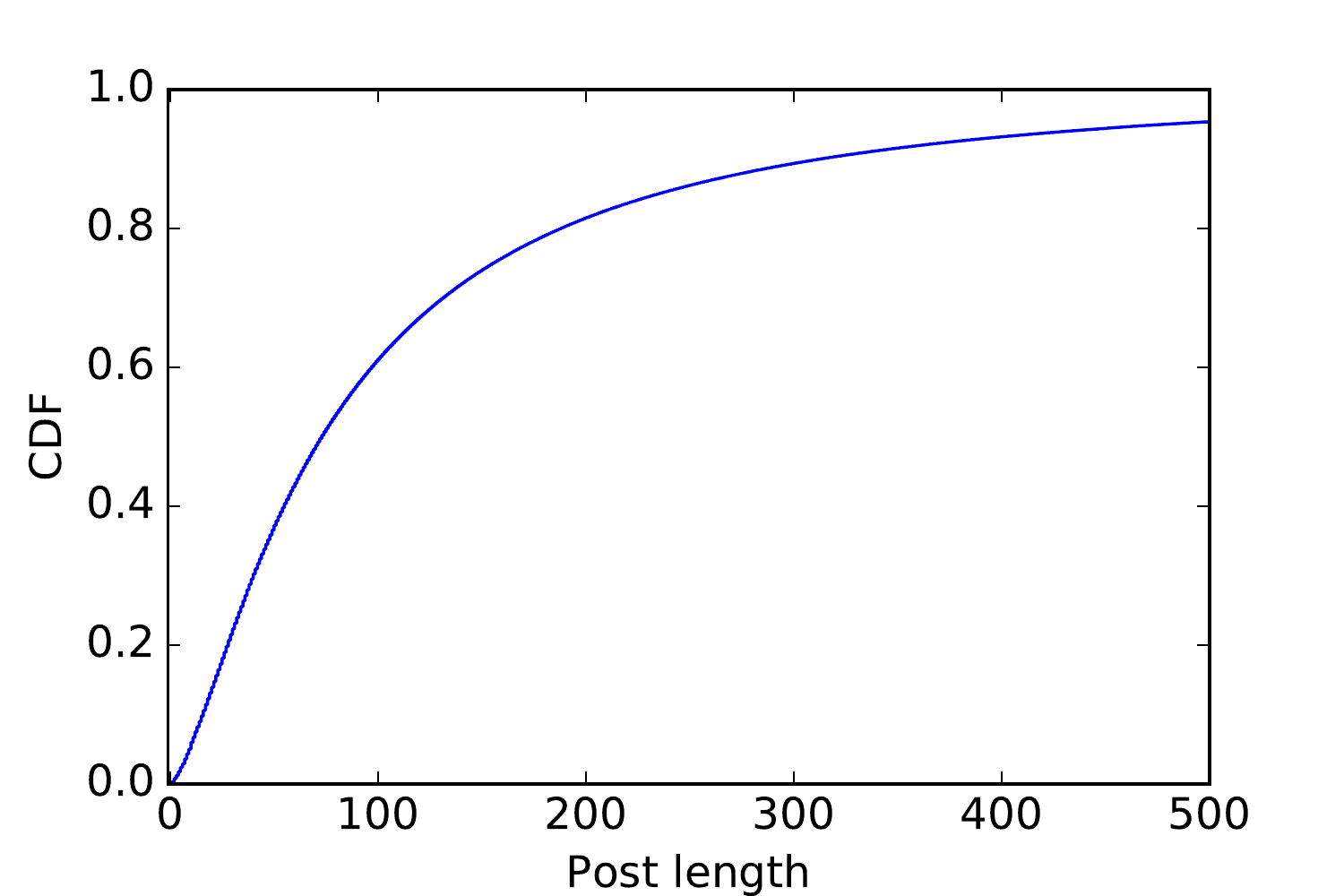}
\caption{}
\label{fig:postlencdf}
\end{center}
\end{subfigure}
\caption{\captionfontsize Empirical cumulative distribution functions (CDF) of the number of posts per user (a) and the post length (b) in the RSDD dataset. }\label{fig:cdf}
\end{figure*}

\subsection{Self-harm risk classification.}
\label{sec:cl16}
We train our methods to label the ReachOut posts and compare them against the top methods from CLPsych '16.
We use the same experimental protocol as was used
in CLPsych '16; our methods were trained on the 947 training posts and evaluated on the remaining 280 testing posts. We used 15\% of the 947 training posts as validation data.

We report results using the same metrics used in CLPsych, which were: the macro-averaged F1 for the $amber$, $red$, and $crisis$ labels (\textit{non-green} posts); the macro-averaged F1 of $ green $ posts vs. $ amber \cup red \cup crisis $ (\textit{flagged} posts); and the macro-averaged F1 of $ green \cup amber $ vs. $ red \cup crisis $ (\textit{urgent} posts).
The \textit{non-green} F1 was used as the official CLPsych metric with the intention of placing emphasis on classification performance for the non-green categories (i.e., those that required some response). The binary \textit{flagged} meta-class was chosen to measure models' abilities to differentiate between posts that require attention and posts that do not, and the binary \textit{urgent} meta-class was chosen to measure their abilities to differentiate between posts that require quick responses and posts that do not. In addition to macro-averaged F1, CLPsych also reported the accuracy for each category. We additionally report F1 macro-averaged over all classes.

\textbf{Results.}
The results on the self-harm risk assessment task for our models and for the current best-performing methods (briefly explained in \S \ref{sec:related}) are shown in Table \ref{table:results1}. We also report a baseline result which is based on a SVM classifier with bigram features. When measured by \textit{non-green} F1, the official metric of the CLPsych '16 Triage Task, our proposed models perform up to 19\% better than the best existing methods. Similarly, our models perform up to 11\% better when measured with an F1 macro-averaged across all categories (i.e., \textit{all} column) and up to 5\% better with measured accuracy across all categories. \modcce performs best in all of these cases, though the difference between the performance of \modcce and
\modclass with an ordinal margin is not statistically significant.

We also evaluate the performance of our methods on the training set using 10-fold cross validation to better observe performance differences (Table \ref{table:results1cv}).
All model variants perform substantially better on the training set than on the test set. This is partially explained by the fact that the models were tuned on the training set, but the large difference in some cases (e.g., the increase in the highest non-green F1 from 0.50 to 0.87) suggest there may be qualitative differences between the datasets. The best-performing method on the test set, \modcce, performs the worst on the training set; worst-performing method on the test set, \modmse, performs the best on the training set. \modclass \textit{(Ordinal)} performs well on both the testing and training sets, however, suggesting that it is more robust than the other methods. Furthermore, there is no statistically significant difference between \modclass \textit{(Ordinal)} and the best-performing method on either dataset.

\section{Analysis}
\subsection{Posts per user and post length}
\label{sec:maxposts}
In this section we consider the effects of the maximum number of posts per user (i.e., $n_{post}$) and the maximum post length
(i.e., $n_{term}$) on the Reddit dataset.
To do so we train the CNN-R model as described in \S \ref{sec:setup} and report F1 on the validation set.
When varying $n_{post}$ we set $n_{term}=100$, and when varying $n_{term}$ we set $n_{post}=1500$.

As shown in Figure \ref{fig:model-params}, the best performance of the CNN-R model is reached when it considers 100 terms in posts and up to 1750 posts for each user. F1 increases as $n_{post}$ increases, up to the maximum tested value of 1750 (Figure \ref{fig:maxposts}).
There is relatively little change in F1 from $n_{post}=1250$ to $n_{post}=1750$, however,
so we use $n_{post}=1500$ in our experiments for efficiency reasons.
As shown in Figure \ref{fig:postcdf}, approximately 20\% of users have more than 1500 posts.
The effect of the maximum post length is not consistent (Figure \ref{fig:maxlen}),
but performance is maximized at $n_{term}=100$. As shown in Figure \ref{fig:postlencdf}, approximately 40\%
of posts are longer than 100 terms.

\begin{table*}[tb]
\begin{center}
\tablefontsize
\renewcommand{\arraystretch}{0.8}
\begin{tabular}{@{}lccccccr@{}}
\toprule
\multirow{2}{*}{Post Selection} & \multicolumn{3}{c}{$n_{posts}=400$} & \multicolumn{3}{c}{$n_{posts}=1500$} & \\
\cmidrule(lr){2-4} \cmidrule(lr){5-7}
& Precision & Recall & F1 & Precision & Recall & F1 \\
  \midrule
  Earliest & \textbf{0.58} & 0.46 & 0.52 & 0.59 & 0.55 & 0.57 \\
  Latest & \textbf{0.58} & 0.50 & \textbf{0.54} & 0.69 & \textbf{0.59} & 0.64 \\
  Random & 0.52 & \textbf{0.53} & 0.53 & \textbf{0.71} & \textbf{0.59} & \textbf{0.65} \\
\bottomrule
\end{tabular}
\caption{\captionfontsize
  Models' performance on RSDD's validation set with different post selection strategies
  and values of $n_{post}$. CNN-E corresponds to the earliest strategy with $n_{post}=400$
  and CNN-R corresponds to the random strategy with $n_{post}=1500$.}
\label{table:selection}
\end{center}
\end{table*}

\begin{table}[tb]
\begin{center}
\tablefontsize
\setlength{\tabcolsep}{8pt}
\renewcommand{\arraystretch}{0.9}
\begin{tabular}{@{}ll@{}}
\toprule
\multicolumn{2}{c}{Top Phrases} \\ \midrule
i went to & to scare you \\
my whole & to have it \\
sometimes i & my son was \\
i'm so sorry & it wasn't \\ \bottomrule
\end{tabular}
\caption{\captionfontsize Example phrases that strongly contributed to a user's depression classification on the RSDD dataset.}
\label{table:analysis}
\end{center}
\end{table}

\subsection{Post selection}
\label{sec:selection}
For users with more than the maximum number of posts $n_{post}$, a post selection strategy dictates
which posts are used as input to the model.
Table \ref{table:selection} shows the effect of the post selection strategy on the Reddit dataset's validation set.
Selecting a user's earliest posts performs the worst
regardless of $n_{post}$'s value, though the differences in F1 are smaller when $n_{post}=400$.
Randomly selecting posts for each user performs the best across all metrics when $n_{post}=1500$,
with a large increase in precision over selecting users' earliest posts
and a small increase over choosing users' latest posts.

\begin{table*}[tb]
\begin{center}
\tablefontsize
\renewcommand{\arraystretch}{0.8}
\begin{tabular}{@{}lccccccc@{}}
\toprule
\multirow{2}{*}{Method} & Non-green & \multicolumn{2}{c}{Flagged} & \multicolumn{2}{c}{Urgent} & \multicolumn{2}{c}{All} \\
\cmidrule(lr){2-2} \cmidrule(lr){3-4} \cmidrule(lr){5-6} \cmidrule(lr){7-8}
 & F1 & F1 & Acc. & F1 & Acc. & F1 & Acc. \\
\midrule
Categorical Cross Ent. & \textbf{0.37} & 0.88 & \textbf{0.90} & 0.48 & 0.83 & \textbf{0.50} & \textbf{0.71} \\
MSE & 0.31 & 0.84 & 0.84 & \textbf{0.54} & \textbf{0.84} & 0.44 & 0.64 \\
Class Metric & 0.30 & 0.88 & 0.89 & 0.46 & 0.81 & 0.45 & 0.68 \\
Class Metric (Ordinal) & 0.34 & \textbf{0.89} & \textbf{0.90} & 0.49 & 0.81 & 0.48 & 0.69 \\
\bottomrule
\end{tabular}
\caption{\captionfontsize Self-harm risk assessment performance on the ReachOut CLPsych '17 test set.
All methods perform substantially worse than on the CLPsych '16 test data.
The difference in performance between the following method pairs are statistically significant (McNemar's test, $p<0.05$): \modcce and \modmse, and \modmse and \modclass \textit{(Ordinal)}.}
\label{table:resultscl17}
\end{center}
\end{table*}

\subsection{Phrases contributing to classification}
In this section we analyze the language that strongly contributed to the identification of depressed users on the Reddit dataset.
Unfortunately, it is impossible to show entire Reddit posts without compromising users' anonymity; we found that even when a post is paraphrased, enough information remains that it can easily be identified using a Web search engine. For example, one Reddit post that strongly contributed to the author's classification as a depressed user contained the mention of a specific type of abuse and several comments vaguely related to this type of abuse. We attempted to paraphrase this post, but found that any paraphrase containing general language related to both the type of abuse and to the user's comments was enough to identify the user. Thus, to protect the anonymity of the users in our dataset, we do not publish posts in any form.

Rather than publishing posts, we identify key phrases in posts from users who were correctly identified as being depressed.
Phrases from eight self-reported depressed users are shown in Table~\ref{table:analysis}; to prevent these phrases from being used to identify users, we retain only the top phrase from each user. These phrases were identified by using the model's convolutional filter weights to identify posts in the validation dataset that are strongly contributing to the model's classification decision, and then using the convolutional filter weights to identify the phrase within each post that most strongly contributed to the post's classification (i.e., had the highest feature values).

In keeping with the design of our dataset, terms related to depression or diagnoses are not present. Instead, the model identifies phrases that often could be associated with a negative sentiment or outlook. For example, ``my whole'' could be part of a negative comment referring to the poster's whole life. It should be noted that the model makes classification decisions based on the occurrence of phrases across many posts by the same user. Though one can imagine how the phrases shown here could be used to convey negative sentiment, the presence of a single such phrase is not sufficient to cause the model to classify a user as depressed.

\subsection{CLPsych '17 shared task}
\label{sec:cl17}
In this section we report results on the 2017 CLPsych Workshop's self-harm risk classification task.\footnote{The 2017 test data was released after the initial version of this manuscript had been completed. An official overview paper for CLPysch '17 is not yet available at the time of writing.}
While CLPsych '17 featured the same self-harm risk classification task as CLPsych '16
(\S \ref{sec:cl16}), new test data was used to conduct the evaluation.
This provides an opportunity to further evaluate our model on the task of self-harm risk assessment and to conduct an error analysis.
The methods were configured and evaluated in the same manner as described in \S \ref{sec:cl16}.\footnote{The results in this section differ slightly from the methods' results as reported by CLPsych '17. Here the methods were trained on only CLPsych '16 training data to match the experimental setup described earlier, whereas the methods were trained on both the CLPsych '16 training and test data in the official results reported by CLPsych '17.}

Results are shown in Table~\ref{table:resultscl17}.
All methods perform substantially worse than they performed on the CLPsych '16 test data as measured by non-green, urgent, and overall F1.
The trends across methods remain similar, however, with \modcce performing the best as measured by non-green and overall F1, and with no statistically significant difference between \modclass \textit{(Ordinal)} and the best performing method.

Notably, the methods' flagged F1 scores do not see a similar decrease on the CLPsych '17 data.
This suggests that the decreased performance is being caused by an inability to distinguish between the
non-green classes (i.e., amber, red, and crisis).
The importance of differentiating between the red and crisis classes increased
with the 2017 shared task, because the proportion of crisis labels in the data increased from
0.4\% (2016 testing) and 4\% (2016 training) to 11\% (2017 testing).
The methods rarely classify a post as crisis, however, causing an increase in the number
of misclassifications on the 2017 testing data.
For example, \modclass \textit{(Ordinal)} classified only four posts from the 2017 test data as crisis,
and it classified no posts from the 2016 test data as crisis.
We leave improving the model to better identify crisis posts as future work.

\balance
\section{Conclusion}
In this work, we argued for the close connection between social media and mental health, and
described a neural network architecture for performing self-harm risk classification and depression detection on social media posts.
We described the construction of the Reddit Self-reported Depression Diagnosis (RSDD)
dataset\footnote{\url{http://ir.cs.georgetown.edu/data/reddit\_depression/}},
containing over 9,000 users with self-reported depression diagnoses
matched with over 107,000 similar control users; the dataset is available under a data usage agreement.
We applied our classification approach to the task of identifying depressed
users on this dataset and found that it substantially outperformed strong existing methods in terms of Recall and F1.
While these depression detection results are encouraging, the absolute values of the metrics illustrate
that this is a challenging task and worthy of further exploration.
We also applied our classification approach
to the task of estimating the self-harm risk posed by posts on the ReachOut.com mental health support forum, and found
that it substantially outperformed strong previously-proposed methods.

Our approach and results are significant from several perspectives:
they provide a strong approach to identifying posts indicating a risk of self-harm in social media;
they demonstrate a means for large scale public mental health studies surrounding the state of depression; and they demonstrate the possibility of sensitive applications in the context of clinical care, where clinicians could be notified if the activities of their patients suggest they are at risk of self-harm.
Furthermore, large-scale datasets such as the one presented in this paper can provide complementary information to existing data on mental health which are generally relatively smaller collections.

\bibliography{emnlp2017}
\bibliographystyle{emnlp_natbib}

\end{document}